\documentclass[aps,amsmath,amssymb,letter,scriptaddress,prl,showkeys]{revtex4}

	\usepackage{amsmath}
	\usepackage{amsthm}
	
	\usepackage{amssymb}
	\usepackage{pgfplots}
	\usepgfplotslibrary{groupplots}
	\usepackage{mathtools}
	\usepackage{makeidx}
	\usepackage{amsfonts}
	\usepackage[ansinew]{inputenc}
	\usepackage[usenames,dvipsnames]{pstricks}
	\usepackage{subfigure}
	\usepackage{epsfig}
	\usepackage{pst-grad} 
	\usepackage{pst-plot} 
	\usepackage[colorlinks,hyperindex]{hyperref}
	\usepackage{algorithm}
    \usepackage{algpseudocode}
	\makeatletter

\newcommand{\Rmnum}[1]{\expandafter\@slowromancap\romannumeral #1@}
\makeatother

\makeindex

\begin{document}

\title{Modeling natural language emergence with integral transform theory and reinforcement learning}

\author{Bohdan B. Khomtchouk}
\email{bohdan@stanford.edu}
\affiliation{Stanford University \\ Department of Biology \\ Stanford, CA 94305, USA}

\author{Shyam Sudhakaran}
\email{shyamsnair@protonmail.com}
\affiliation{Quiltomics \\ Palo Alto, CA 94306, USA \\}



\begin{abstract}
Zipf's law predicts a power-law relationship between word rank and frequency in language communication systems and has been widely reported in a variety of natural language processing applications. However, the emergence of natural language is often modeled as a function of bias between speaker and listener interests, which lacks a direct way of relating information-theoretic bias to Zipfian rank. A function of bias also serves as an unintuitive interpretation of the communicative effort exchanged between a speaker and a listener. We counter these shortcomings by proposing a novel integral transform and kernel for mapping communicative bias functions to corresponding word frequency-rank representations at any arbitrary phase transition point, resulting in a direct way to link communicative effort (modeled by speaker/listener bias) to specific vocabulary used (represented by word rank). We demonstrate the practical utility of our integral transform by showing how a change from bias to rank results in greater accuracy and performance at an image classification task for assigning word labels to images randomly subsampled from CIFAR10. We model this task as a reinforcement learning game between a speaker and listener and compare the relative impact of bias and Zipfian word rank on communicative performance (and accuracy) between the two agents.



\end{abstract}


\maketitle

\section{Introduction and Background}

The linguist George Kingsley Zipf made the observation that the frequency of a word is proportional to the inverse of the word's rank in a text. If the most common word occurs at frequency $n$, then the second most common word occurs at frequency $n/2$, the word with rank three at frequency $n/3$, etc. Generalized, Zipf's law \cite{zipf1949} states:

\begin{equation}
f \propto \frac{1}{r^\alpha}
\label{zipf}
\end{equation}
where $r$ is the word rank and $f$ the frequency in the text, and $\alpha$ is the scaling coefficient generally found to be near 1.0 for many of the texts examined \cite{ferrer_pnas, ferrer_2013, alday_2016, moreno_2016}.

\begin{figure}
\centerline{\includegraphics[width=\textwidth]{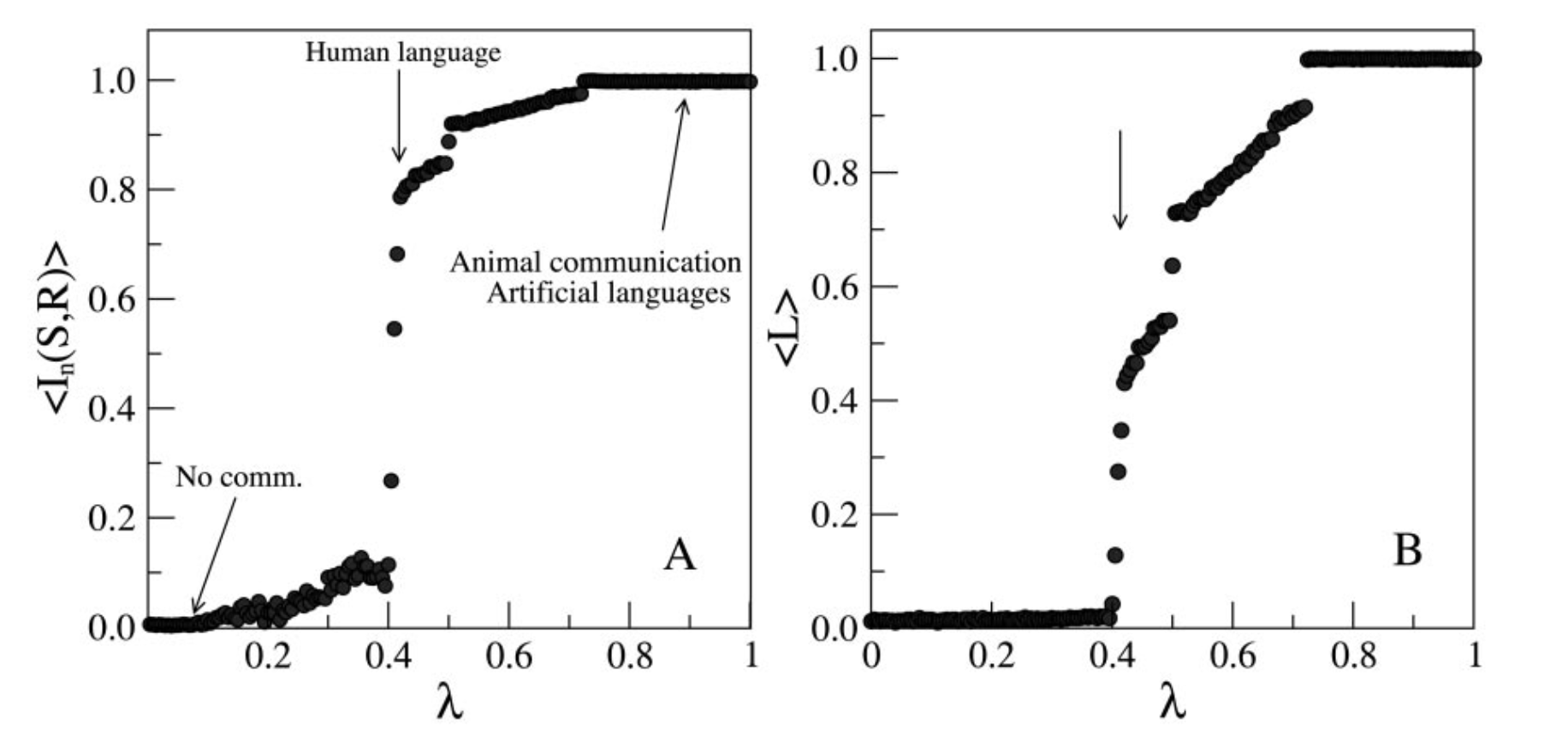}}
\caption{Phase transition in the mutual information $\langle I_n(S, R) \rangle$ and lexical size $\langle L\rangle$ of simulated languages as a function of the proportion of effort, i.e., bias ($\lambda$), devoted to listener interests as opposed to speaker interests. Reproduced with permission from Ferrer i Cancho and Sol\'e (2003).}\label{ferrer}
\end{figure}

Ferrer i Cancho's research group formalized the least-effort principle as it applies to Zipf's law \cite{ferrer2003, ferrer_pnas, ferrer2007, ferrer2010} by employing a mutation-driven genetic algorithm. Here the listener and speaker have different and conflicting interests. The listener seeks to gain as much information as possible from a communicative exchange, and would benefit if there were no ambiguity between word-object mappings. This is the case in which the correlation between words and objects is highest; in information theory \cite{shannon}, this corresponds to a high mutual information, or $I(S,R)$ where $S$ represents the symbol and $R$ the referent or object. The speaker on the other hand looks to minimize her effort in communicating and would benefit from fewer words to choose from, assuming that the choice of words comes with an effort; in information theory, this is quantified using information entropy or $H(S)$.  To this end, Ferrer i Cancho \cite{ferrer_pnas} introduced an energy function based on information theory that models the speaker's and listener's interests:

\begin{equation}
\Omega(\lambda) =  \lambda I(S,R) - (1-\lambda)H(S)
\label{energy}
\end{equation}
where $\lambda$ (0 $\textless$ $\lambda$ $\textless$ 1) controls the balance between the speaker interests, $H(S)$, and listener interests, $I(S,R)$.  It is found \cite{ferrer2003, ferrer_pnas, ferrer2007} that natural languages emerge at the phase transition (Fig.\,\ref{ferrer}) near $\lambda^{*} \approx 0.5$ (i.e., when listener and speaker interests are weighted about equally).  For $\lambda < \lambda^{*}$, there is little or no communication because there are few words in the lexicon $\langle L\rangle$ (Fig.\,\ref{ferrer}B) while, it is assumed, the number of objects remains constant which produces tremendous ambiguity in word-meaning mappings -- one or a few words point to all the objects (i.e., low $I(S,R)$, (Fig.\,\ref{ferrer}A)).  For $\lambda > \lambda^{*}$, there is extremely efficient communication involving single word-single object mappings (i.e., high $I(S,R)$) -- though this comes at a high cost for the speaker (i.e., high $H(S)$) because the lexicon abruptly rises to the number of objects.

\begin{figure*}[ht]
\begin{center}
\centerline{\includegraphics[width=\textwidth]{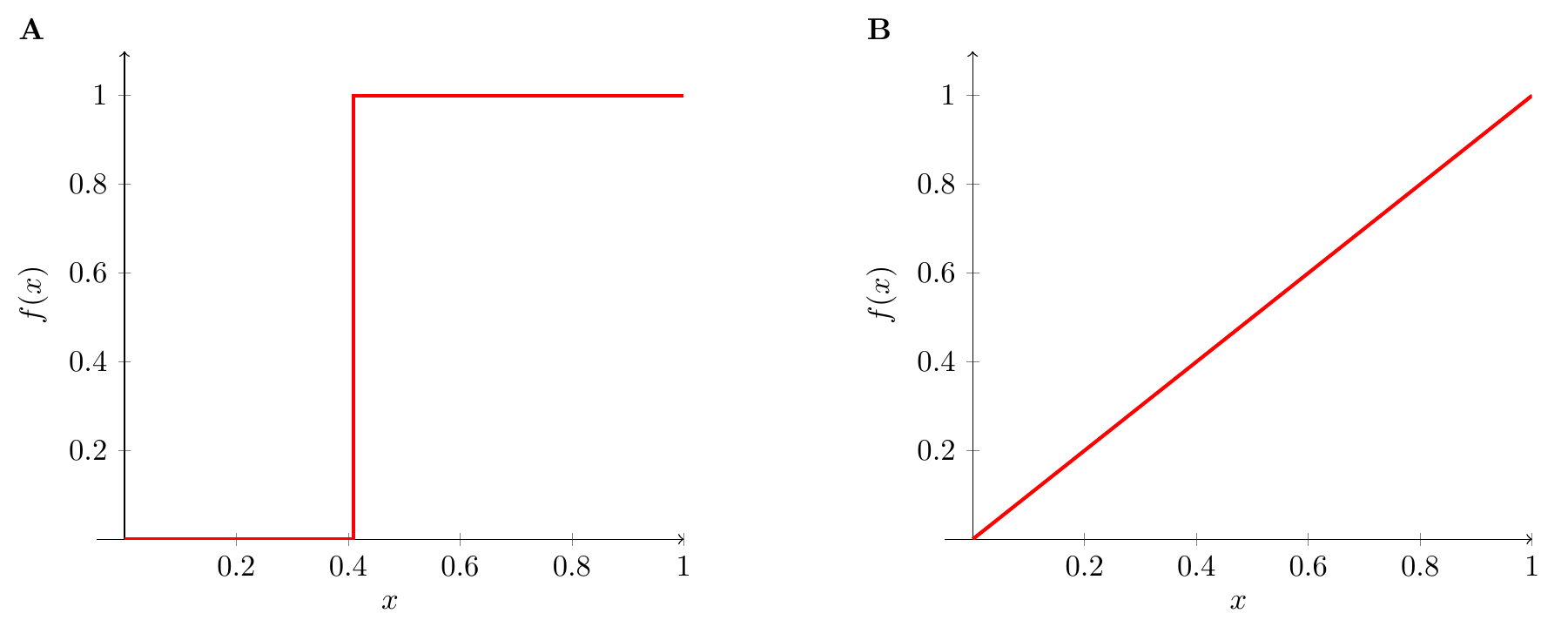}}
\caption{(A) The unit step (Heaviside function) with phase transition at $\lambda=0.41$. (B) The unit ramp function on domain $[0,1]$.}\label{afoto2}
\end{center}
\end{figure*}

The form of both of these phase transitions (Fig.\,\ref{ferrer}) lies somewhere between a step (Heaviside) function and a ramp function (Fig.\,\ref{afoto2}).  The unit ramp function increases gradually, one unit per unit time.  The abrupt switching between states [$x<0$, $f(x)=0$; $x>0$, $f(x)=1$] is typical of electrical circuits \cite{spiegel} and neural systems \cite{mcculloch}.  Indeed, prior studies performed analytical derivations of global minima from equation \eqref{energy} to prove that this theoretical phase transition is well modeled by a step function \cite{ferrer2007, prokopenko}.  These studies demonstrated that the domain $\lambda < \lambda^{*}$ is characterized by single-signal systems (i.e., one signal refers to all objects), the domain $\lambda = \lambda^{*}$ is characterized by non-synonymous systems (i.e., no two signals refer to the same object, although one signal may refer to multiple objects), and the domain $\lambda > \lambda^{*}$ is characterized by one-to-one mappings between signals and objects.  

In mathematics, a transform is a method used to convert an equation in one variable to an equation in a different variable \cite{korner}. Integrals are a common type of transform and have the generalized form:

\begin{equation}
T[f(x)] = F(z) = \int_a^b \! f(x)g(x,z) \, \mathrm{d}x
\label{transform}
\end{equation}
where $f(x)$ is the function being transformed, $T$ is the generalized mathematical transform, and $g(x,z)$ is the kernel of the transform. When the definite integral is evaluated, the variable $x$ drops out of the equation and one is left with a function purely of $z$.  For example, in a Laplace transform \cite{spiegel}, the kernel is the negative exponential $e^{-xz}$, which serves as a damping function. In the special case that $f(x)$ is the unit step function (Fig.\,\ref{afoto2}A), the Laplace transform simply yields $1/z$. For example, in electrical engineering, the Laplace transform is often used to map the behavior of functions in the time domain, $f(t)$, to the frequency domain, $F(z)$.

\section{Results}

We propose a new integral transform called the Slavi transform, $\mathcal{S}$, to map communicative bias functions to corresponding word frequencies.  Consider the function to transform as $N(\lambda)$: the lexical size $\langle L\rangle$ of a language (i.e., the number of words in the language that are connected and have non-zero probability) as a function of the bias, $\lambda$, imparted to the listener over the speaker (Fig.\,\ref{ferrer}B). Because the lexicon size and word-meaning mappings abruptly change at the phase transition near $\lambda^{*}$ (Fig.\,\ref{ferrer}A,B), we can substitute the unit step function (Fig.\,\ref{afoto2}A) for $N(\lambda)$:

\begin{equation}
\begin{split}
\mathcal{S}[N(\lambda)] = \int_0^1 \! N(\lambda)e^{-\lambda r} \, \mathrm{d}\lambda =\int_0^x \! N(\lambda)e^{-\lambda r} \, \mathrm{d}\lambda + \int_x^1 \! N(\lambda)e^{-\lambda r} \, \mathrm{d}\lambda = \\ \int_0^x \! (0)e^{-\lambda r} \, \mathrm{d}\lambda + \int_x^1 \! (1)e^{-\lambda r} \, \mathrm{d}\lambda = \frac{1}{r}(e^{-xr}-e^{-r}) = N(r, x)
\end{split}
\label{transform}
\end{equation}

\noindent where $x \in [0,1]$ represents the phase transition near $\lambda^{*}$.  Kernels of integral transforms of this form are called Slavi kernels, $e^{-\lambda r}$. Since $e^{-xr}-e^{-r} < 1$ for all $x \in [0,1)$, it follows that:

\begin{equation}
N(r, x) < \frac{1}{r}
\end{equation}

We'll see the importance of this result in Equation \ref{inequality}.  For now, we emphasize four key points to give some intuition behind the utility of the proposed kernel ($e^{-\lambda r}$) used during the mapping: 
\begin{itemize}
\item Lexical size of a language has been transformed from a function of bias ($\lambda$) to a function of rank ($r$).
\item $\lambda > \lambda^{*}$ (one-to-one mappings between signals and objects) corresponds to high $r$ (rare, unique signals specific to one object).  $\lambda < \lambda^{*}$ (single-signal systems where one signal refers to all objects) corresponds to low $r$ (frequent, repetitive words referring to multiple objects).
\item The y-axis is preserved under the transformation: it is still the number of words in the language (i.e., frequency).
\item Applying dimensional analysis validates the prerequisite of dimensionless products, as the product $-\lambda r$ is dimensionless since $\lambda$ is a constant in the range $[0,1]$ (Fig.\,\ref{ferrer}) and $r$ is a rank ($r \in \mathbb{N}$) corresponding to a specific word (i.e., signal) in the lexicon.
\end{itemize}

Investigating the other boundary (Fig.\,\ref{afoto2}B) by substituting the unit ramp function for $N(\lambda)$ and performing the Slavi transform yields $1/r^2$ for $r \to \infty$, a hallmark of complex languages possessing many words (where high $r$ corresponds to rare words in the lexicon):

\begin{equation}
\begin{split}
\mathcal{S}[N(\lambda)] = \int_0^1 \! N(\lambda)e^{-\lambda r} \, \mathrm{d}\lambda =\int_0^1 \! (\lambda)e^{-\lambda r} \, \mathrm{d}\lambda = -\frac{2}{re^{r}} + \frac{1}{r^2} = N(r)
\end{split}
\label{transform2}
\end{equation}

Thus, depending on how abrupt the phase transition is, one should expect most words in a complex language to scale within the range:

\begin{equation}
\frac{1}{r^2} \leq N(r) \leq \frac{1}{r}
\label{inequality}
\end{equation}
or, in terms of the Zipfian exponent, $1 \leq \alpha \leq 2$, which is typically found to be the case \cite{ferrer_last, moreno_2016}.  Taken together, there is a connection between the rank of the $r\textsuperscript{th}$ word and its frequency in the lexicon, $N(r, x)$, provided the language is organizing around a phase transition in mutual information and lexicon size. 

\section{Simulation}
We want to further demonstrate the importance of the Slavi transform mapping the lexical size of a language, originally a function of bias, to a function of rank by introducing a reinforcement learning simulation. Our goal for this simulation is to introduce a realistic scenario where we can show that word rank is more impactful than bias for tasks involving communication performance. Higher word rank corresponds to more infrequent and unique words, while lower word rank corresponds to more frequent words and synonyms. To better understand the inspiration behind the simulation, consider the following scenario:

You are walking with your friend in a very busy park. There are numerous objects that you and your friend see, but a brown colored dog captures your attention and you want to communicate this to your friend verbally. In this case, you are the ``speaker'' and your friend is the ``listener''. You, the speaker, generate a ``label'', such as ``brown dog'', to alert the listener's attention to the dog in question. ``Brown dog'', is a very specific label and therefore it makes it very easy for the listener to associate it with the specific dog the speaker intends to communicate. The phrase, ``brown dog'', is exclusively for a dog whose color is brown, which means that the phrase is mapped to the brown colored dog or similar objects like it in the environment. In this scenario, the speaker is exerting more effort than the listener, who can easily map ``brown dog'' to the brown colored dog. If the speaker instead just says ``brown thing'', the listener will have to exert more effort to figure out that the speaker is talking about the dog (and, therefore, map its location in the environment). This would mean that the listener is exerting more effort than the speaker. 

Recall from Eq.\,\ref{energy}  that this effort measurement is represented as bias in Ferrer i Cancho's energy function.  Bias (i.e., $\lambda$) controls the balance between the speaker interests, $H(S)$, and listener interests, $I(S,R)$. It can be viewed as a learning rate, which will help with the formulation of the training model later on. It is also important to note that even if the listener initially has no idea that the ``brown dog'' means brown colored dog, through repetition it would learn this association, in order to communicate effectively with the speaker. This can be viewed as natural language emergence, where labels are chosen (or ``generated") and repeated until the listener and speaker reach a consensus. Our main goal is to simulate this natural language emergence and compare the effects of bias and word rank on the ability to communicate efficiently between a speaker and a listener, ultimately testing the hypothesis that the Slavi transform (i.e., transforming bias to rank) is useful for practical computer vision tasks such as image classification.  

\section{Data and source code}
We chose a set of 10 unique image classes, provided by the CIFAR10 dataset. The image classes are ``airplane'', ``automobile'', ``bird'', ``cat'', ``deer'', ``dog'', ``frog'', ``horse'', ``ship'', and ``truck''.  Source code accompanying this simulation can be found here: \href{https://github.com/Quiltomics/NLERL}{https://github.com/Quiltomics/NLERL}

\section{Model Structure}
We attempt to model the scenario introduced above with a two-player game between two reinforcement learning agents, a speaker and a listener, where images are the objects that are to be communicated. The speaker and listener start off as almost independent, not communicating successfully. Through a training process they will create a language, or a mapping, between objects and one hot encoded values to communicate effectively. We adapted the model introduced by Angeliki Lazaridou, Alexander Peysakhovich, Marco Baroni \cite{multi_agent}.

The game between the speaker, parameterized as $\theta_{s}$, and listener, parameterized as $\theta_l$, is as follows: 
\begin{enumerate}
    \item A sample image from each of the $n$ unique classes from an image dataset is drawn and passed through a pretrained VGG19 network \cite{vgg19}, the output represented by image vectors  $\left\{i_0,...,i_{n-1}\right\}$. One of the vectors is chosen to be the ``target image'', represented as $i_t \in \left\{i_0,...,i_{n-1}\right\}$, where $t\in\left\{0,...,n-1\right\}$.
    
    \item The speaker takes as input the target image $i_t$ and generates a label from a vocabulary of size $m$, where $m > 1$. The label is represented as a one hot encoded vector of size $m$. This action is the speaker's policy, $\pi_{\theta_{s}}(i_t,m)$.
    
    \item  The listener takes in each image vector, $\left\{i_0,...,i_{n-1}\right\}$, and the action label, $\pi_{\theta_{s}}(i_t,m)$, generated by the speaker. It tries to guess which image the speaker saw by matching the label generated by the speaker to the correct target image $i_t$. This guess is the listener's policy, $\pi_{\theta_{l}}(\left\{i_0,...i_{n-1}\right\}, \pi_{\theta_{s}}(i_t,m))$.
    
    \item If the listener guesses the target correctly, or $\pi_{\theta_{l}}(\left\{i_0,...i_{n-1}\right\}, \pi_{\theta_{s}}(i_t,m)) = i_t$, then both the speaker and listener receive a reward of 1. If the listener gets it wrong, the speaker and listener receive a reward of 0.
    
    \item Update parameters $\theta_{s}$ and $\theta_{l}$.
\end{enumerate}
Over time, the listener and the speaker will develop a mapping to communicate the target images.
\section{Training and Testing}
We chose the speaker and listener to be reinforcement learning agents because the way they learn to communicate effectively is similar to how humans would learn: through repetition and based on whether the speaker and listener reached a consensus or not. The update rule we chose to optimize the speaker's and listener's parameters is based off the Monte Carlo Policy Gradient (REINFORCE) algorithm \cite{reinforce}:
\begin{algorithm}[H]
  \caption{Monte Carlo Policy Gradient (REINFORCE) algorithm }\label{euclid}
  \begin{algorithmic}[0]
    \Procedure{REINFORCE}{}
      \State initialize parameters $\theta$ arbitrarily 
      \For{\texttt{each episode $\left\{s_{0},a_{0},r_{0},...,s_{T},a_{T},r_{T}\right\} \sim \pi_{\theta}$}}
        \For{\texttt{$t = 0$ to $T$}}
          \State generate long term value $v_t$ from function $Q^{\pi}(s,a)$
          \State $\theta\gets \theta + \alpha\nabla_{\theta}\log\pi_{\theta}(s_{t},a_{t})v_{t}$
        \EndFor
      \EndFor
      \State \textbf{return} $\theta$
    \EndProcedure
  \end{algorithmic}
\end{algorithm}
We chose the long term reward $v_t$ for our algorithm to simply be $r_t$, because trials are independent of each other. We also chose to modify the update rule by incorporating bias, $0 < \lambda < 1$. Returning to Ferrer i Cancho's energy function (Eq.\,\ref{energy}), we can view bias (i.e., $\lambda$) as the ``learning rate'', measuring the importance of the speaker's and listener's performances, and updating the model accordingly. This lets us easily incorporate $\lambda$ into our simulation. Since $(1-\lambda)$ scales speaker interests, $H(S)$, and $\lambda$ scales listener interests, $I(S,R)$, we can scale our normal learning rate by $(1-\lambda)$ for the speaker's update and $\lambda$ for the listener's update and formulate a modified Monte Carlo Policy Gradient (REINFORCE) algorithm:
\begin{algorithm}[H]
  \caption{Modified Monte Carlo Policy Gradient (REINFORCE) algorithm }
  \label{euclid}
  \begin{algorithmic}[0]
    \Procedure{Modified REINFORCE}{}
      \State  $s_{t}$ = speaker state at time step $t$ 
      \State  $l_{t}$ = listener state at time step $t$ 
      \State initialize speaker parameters $\theta_{s}$ and listener parameters $\theta_{l}$ arbitrarily. 
      \For{\texttt{each episode $\left\{s_{0},l_{0},\pi_{\theta_{s}}(s_{0}),\pi_{\theta_{l}}(l_{0},\pi_{\theta_{s}}(s_{0})),r_{0},...,s_{T},l_{T},\pi_{\theta_{s}}(s_{T}),\pi_{\theta_{l}}(l_{T},\pi_{\theta_{s}}(s_{T})),r_{T}\right\}$}}
        \For{\texttt{$t = 0$ to $T$}}
          \State update speaker parameters: $\theta_{s}\gets \theta_{s} + (\alpha\times(1-\lambda))\nabla_{\theta_{s}}\log\pi_{\theta_{s}}(s_{t})r_{t}$
          \State update listener parameters: $\theta_{l}\gets \theta_{l} + (\alpha\times\lambda)\nabla_{\theta_{l}}\log\pi_{\theta_{l}}(l_{t},\pi_{\theta_{s}}(s_{t}))r_{t}$
        \EndFor
      \EndFor
      \State \textbf{return} $\theta_{s},\theta_{l}$
    \EndProcedure
  \end{algorithmic}
\end{algorithm}
\section{Agent Architectures}
Both the speaker and listener are feedforward neural networks, implemented in Keras. The neural networks' weights are initialized using Glorot Initialization \cite{glorot}. The speaker takes the target image as input and passes it through a pretrained VGG19 to generate an image vector embedding. The speaker passes the embeddings through its hidden layers to output a softmax probability vector of vocabulary size $m$. The speaker samples an action, or a label, from the generated probability vector. The listener passes a sampled image from each class, including the target image, through a pretrained VGG19 to generate image vector embeddings. The listener takes in the speaker's generated label as an additional input. The image vector embeddings are passed through a shared hidden layer to create a new embedding. The speaker's generated label is passed through a separate hidden layer to create a label embedding. Dot products are computed between each new image embedding and label embedding. A softmax probability vector of size $10$ is generated from these dot products. A comprehensive workflow is illustrated in Fig. 3.   
\begin{figure}
    \includegraphics[width=6cm]{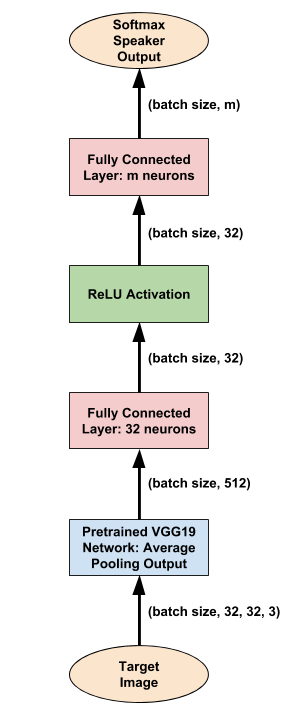}
    \includegraphics[width=11cm]{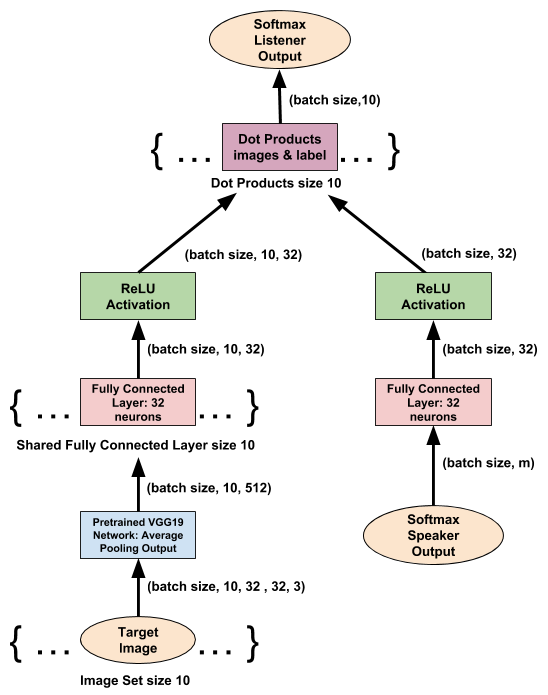}
\caption{(Left) Speaker model architecture: The speaker takes in a random batch of target images. After the images are passed through a pretrained VGG19 network, they are passed through a 32 neuron fully connected layer with ReLU activation to generate a target image embedding. The embedding is passed through a fully connected softmax output layer with the number of neurons equal to the vocabulary size $m$. (Right) Listener model architecture: The listener takes in two inputs -- a batch of samples of each image class, including the target image, and the softmax speaker output of size $m$. The images are passed through a pretrained VGG19 network and then one shared 32 neuron fully connected layer to create image embeddings. The softmax speaker output is passed through a separate 32 neuron fully connected layer with ReLU activation to create a label embedding. The dot products between the label embedding and the image embeddings are computed and passed through a softmax activation.}
\end{figure}
\section{Simulation Results}
To simulate word rank, we simply modify the vocabulary size $m$: smaller vocabulary corresponds to more frequent labels for multiple objects, resulting in lower word rank, while larger vocabulary corresponds to more one-to-one mappings between labels and objects, resulting in higher word rank. This is very intuitive to how humans interact, if there are less words in a speaker's vocabulary than objects, the speaker will be forced to use the same word more frequently to mean different objects, making it harder to communicate. We trained for 1000 episodes, with 100 samples from each image class, and used a learning rate $\alpha$ of 0.001. 

To reiterate, our main goal of the simulation has been to test the practical utility of the Slavi transform, namely to see if transforming bias to rank is useful for machine learning tasks. We wish to test the hypothesis that word rank is a better and more intuitive alternative to bias for effective communication. To do this, we observed the effect on the performance of our model while changing vocabulary size $m$ and keeping bias $\lambda$ constant compared to the performance while changing bias $\lambda$ and keeping vocabulary size $m$ constant. Performance is measured as accuracy, or total reward / total number of trials, which means the proportion of trials where the receiver picked the right object. We then determine whether there is a stronger trend between accuracy and rank compared to accuracy and bias. To test word rank, we ran the simulation with vocabulary sizes $m$ of 2,10,50,100,250,500,650,800,1000, with a constant $\lambda$ of 0.5 (equal effort between speaker and listener). When testing for bias, we used $\lambda$ values of 0.002,0.01,0.05,0.1,0.25,0.5,0.65,0.8,0.99, and a constant vocabulary size  $m$ of 10. The rolling average of the 100 most recent episodes for both the simulations are shown in Fig. 4. The accuracy values for rank and bias are shown in Table I and Table II, respectively.

\begin{figure}
    \includegraphics[width=8.5cm]{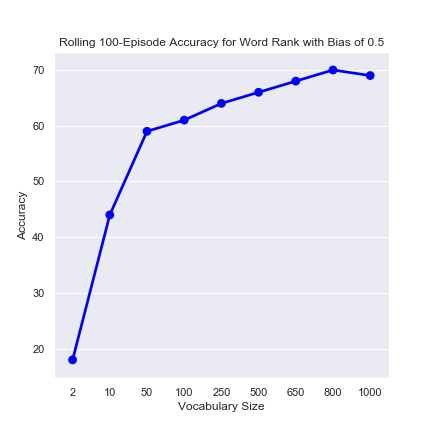}
    \includegraphics[width=8.5cm]{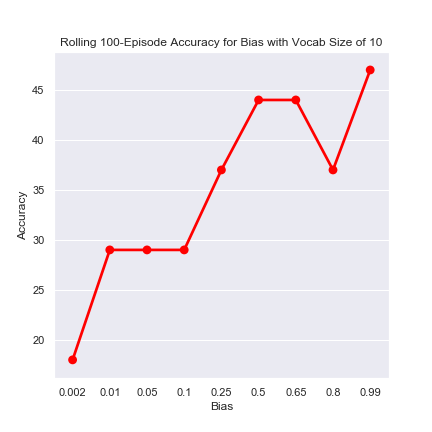}
\caption{(Left) Rolling average accuracy of 100 most recent episodes, over a training period of 1000, of models with different vocabulary sizes and a constant bias ($\lambda$) of 0.5. (Right) Rolling average accuracy of 100 most recent episodes, over a training period of 1000, of models with different bias ($\lambda$) values and a constant vocabulary size of 10.}
\end{figure}

\begin{table}[H]
    \begin{minipage}{.5\linewidth}
        \caption{Rolling Accuracy 100 episodes Word Rank}
        \centering
        \begin{tabular}[t]{|p{3.2cm}|c|}
            \hline
            {\bf Vocabulary Size}&{\bf Accuracy}\\ \hline
            2&18\% \\ \hline
            10&44\% \\ \hline
            50&59\% \\ \hline
            100&61\% \\ \hline
            250&64\% \\ \hline
            500&66\% \\ \hline
            650&68\% \\ \hline
            800&70\% \\ \hline
            1000&69\% \\ \hline
            {\bf Linear Regression Coefficient}&{\bf 95.38}  \\ \hline
        \end{tabular}
    \end{minipage}%
    \begin{minipage}{.5\linewidth}
        \caption{Rolling Accuracy 100 episodes Bias}
        \centering
        \begin{tabular}[t]{|p{3.2cm}|c|}
             \hline
            {\bf Bias \bf ($\lambda$)}&{\bf Accuracy}\\ \hline
            0.002&18\% \\ \hline
            0.01&29\% \\ \hline
            0.05&29\% \\ \hline
            0.1&37\% \\ \hline
            0.25&44\% \\ \hline
            0.5&44\% \\ \hline
            0.65&37\% \\ \hline
            0.8&37\% \\ \hline
            0.99&47\% \\ \hline
            {\bf Linear Regression Coefficient}&{\bf 59.41} \\ \hline
        \end{tabular}
    \end{minipage} 
  \end{table}
From the accuracy measurements, we can see that both bias ($\lambda$) and word rank (vocabulary size $m$) have a positive relationship with accuracy. However, it is clear that bias has a weaker relationship than word rank. For vocabulary sizes $ m \leq 800$, there seemed to be a consistent increase in model performance, reaching a peak of $70\%$, and from $m > 800$ the accuracy seemed to level off at around $69\%$. Bias seemed to have a weaker relationship: at $0.001 \leq \lambda \leq 0.05$ and $0.25 \leq \lambda \leq 0.5$ the performance of the model did not improve. Furthermore, at $0.65 \leq \lambda \leq 0.8$, the performance actually dipped from $44\%$ to $37\%$, but eventually rose to $47\%$. Observing the Linear regression coefficients for vocabulary size vs accuracy and bias vs accuracy, we can see that a proportional increase in vocabulary size has about a $60\%$ larger expected increase in accuracy than bias ($95.38 / 59.41$). Word rank is not only more intuitive to understand (i.e., vocabulary size is much easier to understand than bias), but also has a stronger positive relationship with accuracy, implying that it has a stronger effect, compared to bias, on the communicative performance between a speaker and a listener.
\section{Future directions}
The peak accuracy we reached was $70\%$, so there is definite room for improvement. The model architecture can be expanded upon to improve the performance -- one idea could be to include 1D convolutional layers, which may improve the accuracy. In addition to model improvements, with more computational power, more image samples and training episodes can be played. It would also be interesting to substitute CIFAR10 with the CIFAR100 dataset (i.e., to have more objects to communicate the problem).
\section{Conclusions}

We show that the Slavi transform maps communicative functions of speaker-listener bias directly to word rank. Specifically, we demonstrate that the lexical size of a language can be mapped from a function of bias, $N(\lambda)$, to a function of rank at any arbitrary phase transition point, $N(r, x)$. We provide a practical example in the form of a unique approach to an image classification task, where two reinforcement learning agents (a speaker and a listener) communicate images and labels with each other. When testing the impact of bias and word rank we observed that word rank had a much stronger positive effect on communicative performance (and accuracy) than bias. This suggests that functions of word rank are generally more useful than functions of bias for modeling communicative systems. This study highlights the importance of integral transform theory to understanding and improving information-theoretic models of communicative systems in the context of Zipfian ranks.


\end{document}